\documentclass[runningheads]{llncs}
\usepackage{booktabs}
\usepackage{multirow}
\usepackage{tabularx}
\usepackage{booktabs}
\usepackage{graphicx}
\usepackage{amsmath}
\usepackage{color}
\begin{document}
\title{GelFlow: Self-supervised Learning of Optical Flow for Vision-Based Tactile Sensor Displacement Measurement}
%
%
\author{Zhiyuan Zhang\inst{1}\orcidID{0000-0001-7156-3511} \and
Hua Yang\inst{1}\orcidID{0000-0002-5430-5630} \and
Zhouping Yin\inst{1}}
\authorrunning{Z. Zhang et al.}
%
\institute{Huazhong University of Science and Technology, School of Mechanical Science and Engineering, State Key Laboratory of Digital Manufacturing Equipment and Technology, Wuhan, China
\email{huayang@hust.edu.cn}}
\maketitle              
\begin{abstract}
High-resolution multi-modality information acquired by vision-based tactile sensors can support more dexterous manipulations for robot fingers.
Optical flow is low-level information directly obtained by vision-based tactile sensors, which can be transformed into other modalities like force, geometry and depth.
Current vision-tactile sensors employ optical flow methods from OpenCV to estimate the deformation of markers in gels.
However, these methods need to be more precise for accurately measuring the displacement of markers during large elastic deformation of the gel, as this can significantly impact the accuracy of downstream tasks.
This study proposes a self-supervised optical flow method based on deep learning to achieve high accuracy in displacement measurement for vision-based tactile sensors.
The proposed method employs a coarse-to-fine strategy to handle large deformations by constructing a multi-scale feature pyramid from the input image.
To better deal with the elastic deformation caused by the gel, the Helmholtz velocity decomposition constraint combined with the elastic deformation constraint are adopted to address the distortion rate and area change rate, respectively.
A local flow fusion module is designed to smooth the optical flow, taking into account the prior knowledge of the blurred effect of gel deformation.
We trained the proposed self-supervised network using an open-source dataset and compared it with traditional and deep learning-based optical flow methods.
The results show that the proposed method achieved the highest displacement measurement accuracy, thereby demonstrating its potential for enabling more precise measurement of downstream tasks using vision-based tactile sensors.

\end{abstract}
\section{Introduction}
Vision and tactile are crucial sources of information for perceiving and interacting with the world \cite{introduction_1}.
With computer vision and robotics advancement, vision-based tactile sensors fusing both modalities are becoming increasingly popular for enabling intelligent robots to perceive and manipulate delicate objects precisely.
A typical visual-tactile sensor hardware comprises three components: a contact module, a camera module, and an illumination module \cite{introduction_3}.
The contact module requires resilient and optically favorable materials, as its performance directly affects the accuracy of subsequent optical measurements.
It is often embedded with a marker layer to visualize the contact material's deformation.
The camera and illumination modules can be classified into two main categories based on the measurement principle: a monocular camera system with multi-color illumination systems and multi-camera systems with a monochromatic illumination system.
The integration of these three modules allows vision-based tactile sensors to capture and measure various types of information, including force \cite{GelForce}, geometry reconstruction \cite{geometry}, sliding detection \cite{slip}, and object recognition \cite{introduction_1}.

The displacement of markers in the contact module provides valuable information for measuring additional physical properties, such as the shape of the contacting object, the forces applied, and the roughness of its surface.
This information can be analyzed by the robot for downstream tasks.
Accurate and dense displacement measurements improve the resolution of other modal information, providing better input for subsequent tasks, thereby enhancing the accuracy of robot perception and manipulation \cite{random_color_pattern}.
However, the contact module composed of gel material is characterized by large elastic deformation, which can result in errors in estimating the displacement of existing vision-based tactile sensors using the optical flow algorithm in OpenCV \cite{OpenCV}.
These inaccuracies in displacement measurements can lead to inaccuracies in the final estimated physical information \cite{Deltact}.
Therefore, our motivation is to develop an accurate pixel-level optical flow estimation method to better deal with the deformation properties of the gel materials.

In this study, we introduce a self-supervised learning optical flow approach, named GelFlow, for a more precise measurement of displacement in gel-like materials.
Our proposed method offers two novel loss terms, namely the Helmholtz velocity decomposition constraint and elastic deformation constraint, and a practical local flow fusion module to track the movement of gel materials captured by a monocular camera.
These contributions improve displacement measurement accuracy and enhance vision-based tactile sensors' capability to estimate physical information.
The rest of this paper is organized as follows.
Section \ref{sec2} provides an introduction to previous works related to vision-based tactile sensors and dense displacement processing.
In Section \ref{sec3}, the structure and individual modules of the proposed GelFlow method are elaborated on and discussed in detail.
The comparison results with other optical flow methods and the ablation study are presented in Section \ref{sec4}.
Finally, the conclusions of this work are discussed in Section \ref{sec5}.
\section{Related Work}\label{sec2}
The ability to perceive and model the contact surface's three-dimensional (3D) geometry is a fundamental feature that distinguishes vision-based tactile sensors from conventional tactile sensors.
Based on different principles of 3D surface reconstruction, vision-based tactile sensors can be divided into two categories: sensors based on photometric stereo reconstruction and sensors based on multi-view geometry reconstruction.
Among the first type of sensors, the GelSight \cite{GelSight} sensor uses an RGB trichromatic light source to illuminate the contact layer and a monocular camera to capture the image.
This algorithm enables it to obtain the normal gradient of each pixel, resulting in high accuracy for contact geometry.
However, this method requires rigorous structural design and light source calibration.
The GelSlim \cite{GelSlim} sensor improves GelSight's optical path system by using a mirror-reflective structure so that the camera no longer has to face the contact body, making the entire sensor compact.
The DIGIT \cite{DIGIT} sensor is low-cost, compact, and provides high-resolution tactile perception, making it more practical for robotic finger manipulation.
Unlike the previous flat structure contact layer, DenseTact \cite{DenseTact} sensor uses a spherical contact layer, making it more suitable for sensing complex object surfaces.
Among the second type of sensors, OmniTact \cite{OmniTact} sensor uses multiple micro-cameras to capture multi-directional high-resolution deformation information to obtain accurate and reliable measurement results.
GelStereo \cite{GelStereo} sensor simplifies the number of cameras required by using binocular cameras to calculate the depth information of the contact surface through the disparity map in left and right views.
Tac3D \cite{Tac3D} further simplifies the number of cameras by using a monocular camera with a mirror system to achieve a pseudo-binocular imaging effect, achieving the same 3D reconstruction purpose.

In addition to 3D reconstruction, other valuable information, such as force estimation and sliding detection, is also crucial for robot perception.
This information is obtained by measuring the deformation of the contact layer and then converting it according to specific criteria.
Optical flow is an essential technique for deformation estimation, and accurate optical flow estimation with high resolution can provide more detailed information for more precise and dexterous operations.
There are two primary approaches to enhancing the reliability of optical flow estimation: utilizing more precise optical flow algorithms and designing better marker layers.
During the early stages of vision-based tactile sensor development, the Lucas-Kanada optical flow method \cite{LK} was utilized to track the movement of markers, which could only produce a sparse vector field.
Interpolation methods are used for upsampling the sparse field, and significant errors occur during this process \cite{random_color_pattern}.
Subsequently, more robust and accurate pixel-level optical flow methods such as Farneback \cite{FB} and Dense Inverse Search (DIS) \cite{DIS} methods were adopted to avoid the interpolation error and improve estimation precision \cite{random_color_pattern}.
The conventional marker layer consists of an array of sparse black dot markers, which does not provide rich deformation information.
Moreover, a single color pattern lacked robustness in optical flow estimation due to the brightness conservation hypothesis.
When using a single color pattern, the similarity of pixels made the optical flow estimation confusing.
In order to overcome the limitations mentioned above, researchers have proposed various types of marker layers.
\cite{Soft_bubble} added high-density internal markers, which enabled dense optical flow tracking for estimating shear-induced membrane displacement.
\cite{random_color_pattern} replaced the sparse black dot markers with a novel random color pattern, which achieved more accurate and higher resolution two-dimensional (2D) deformation estimation.

In this work, our purpose is to propose a pixel-level optical flow method with high accuracy and robustness.
Our method takes advantage of the powerful tools of deep learning, and we hope it can be helpful in deformation estimation and other downstream tasks in vision-based tactile sensors.
\section{Method}\label{sec3}
\subsection{Network Architecture}
\begin{figure*}[ht]
  \centering
  \includegraphics[width=0.9\textwidth,clip]{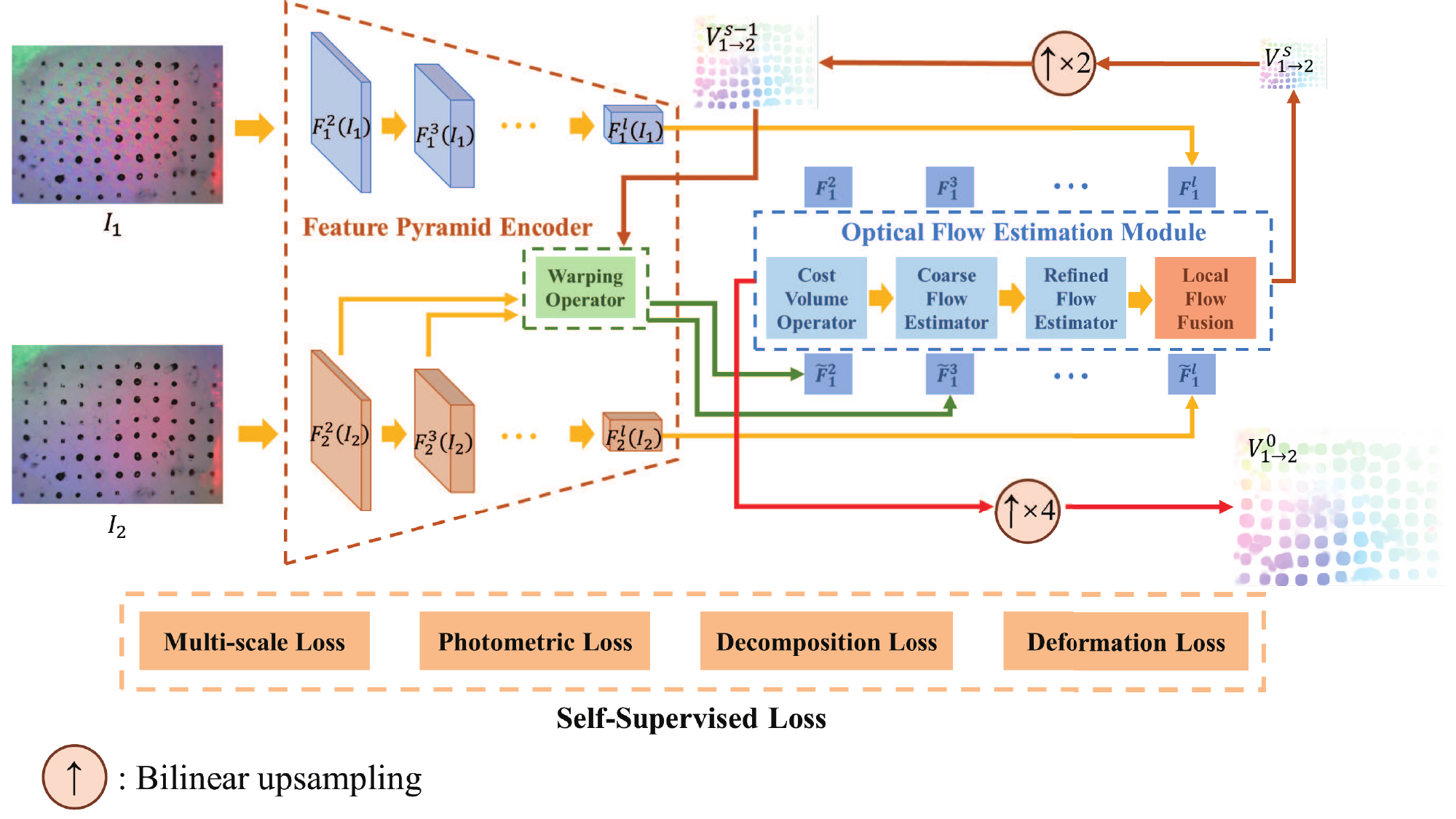}\\
  \caption{GelFlow, the proposed architecture for deformation measurement for gel-like materials using coarse-to-fine strategy, local flow fusion module and Self-supervised losses.}\label{fig:Architecture}
\end{figure*}

Fig. \ref{fig:Architecture} shows the framework of the GelFlow.
First, the encoder module of PWC-Net \cite{sun2018pwc} is adopted to extracted the multi-scale features for the input image pairs $I_1$ and $I_2$ and denoted as $F^{s}_{1}(I_1)$ and $F^{s}_{2}(I_2)$, where the superscript $s$ represents the $s$th scale of the pyramid.
In Optical Flow Estimation Module, apart from basic operators in PWC-Net like Cost Volume Operator, Coarse Flow Estimator and Refined Flow Estimator, we add Local Flow Fusion Module (LFFM) for better gel-like materials deformation estimation.
The output flow at the current pyramid scale $V^{s}_{1{\rightarrow}2}$ is upsampled by a factor of 2 using bilinear interpolation and then used as the initial flow for the subsequent scale $V^{s-1}_{1{\rightarrow}2}$.
According to the coarse-to-fine strategy, the feature of the second image $F^{s}_{2}(I_2)$ are warped \cite{warping} using the output flow (denoted by $\tilde{F}^{s}_{2}(I_2)$) to reduce the feature distance between the extracted feature of the first image $F^{s}_{1}(I_1)$, enabling better handling of large displacement motion.
Note that the input flow at the top scale is set to $\mathbf{0}$, and the final output flow of GelFlow $V^{0}_{1{\rightarrow}2}$ is obtained by bilinear upsampling of the output flow by a factor of 4 at the bottom scale.
The multi-scale strategy allows for the extraction of richer feature information by convolving the input image at different scales, improving the robustness of optical flow estimation.
\begin{figure*}[ht]
  \centering
  \includegraphics[width=0.90\textwidth,clip]{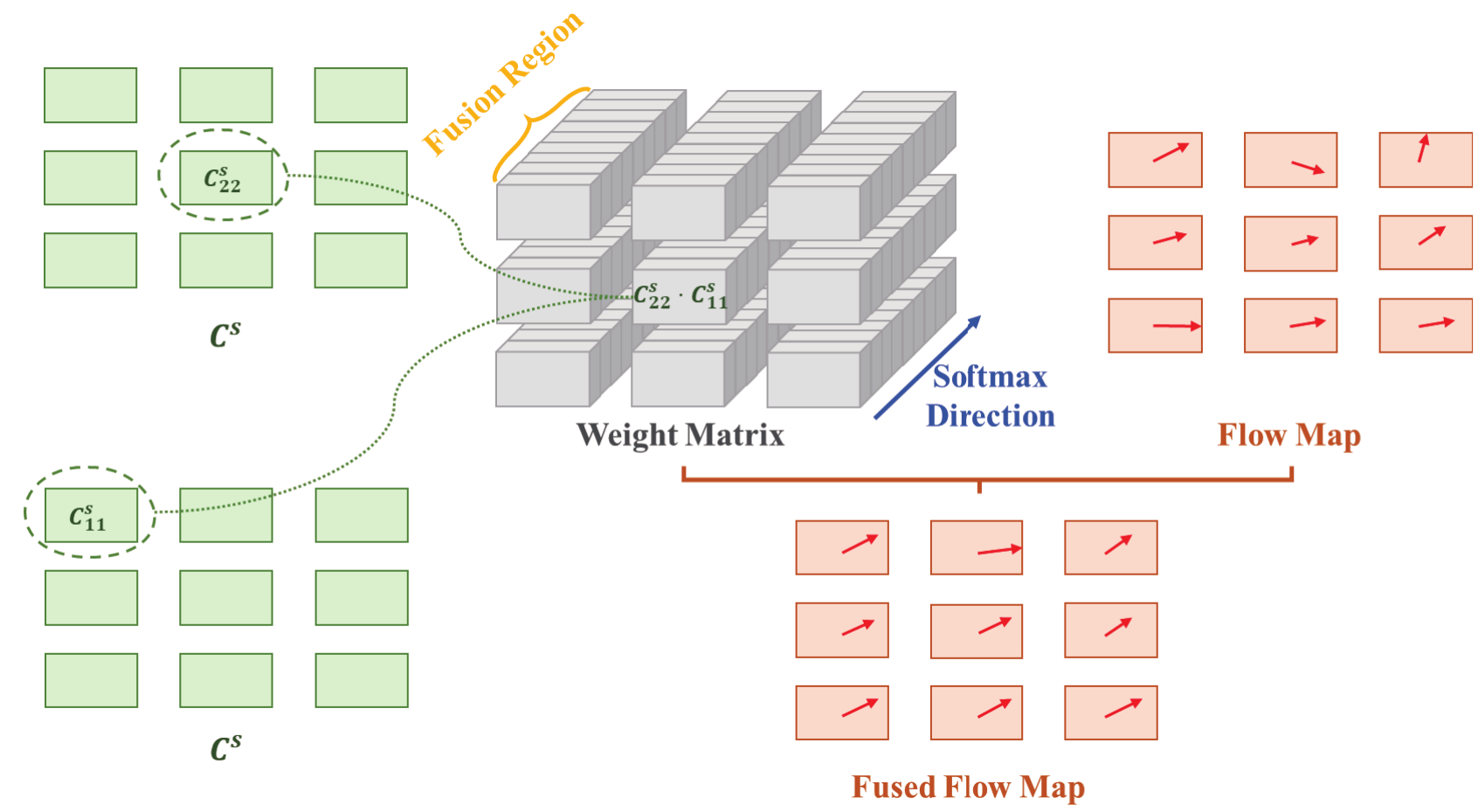}\\
  \caption{Structure of Local Flow Fusion Module. The feature vector at each position is dot producted with the feature vectors in the surrounding area. Then, all of the results within the local range are passed through a softmax function and used as weights for weighted fusion with the estimated optical flow. This process results in a locally smooth optical flow field.}\label{fig:Fusion}
\end{figure*}
\subsection{Local Flow Fusion Module}
As the cross-linking network structure of the gel material results in its deformation showing overall smoothness \cite{TactileFlow}, thus the generated optical flow field exhibits a blurred effect.
Taking advantage of this property, we designed a local optical flow fusion module, and Fig. \ref{fig:Fusion} shows the implementation details of this module.
Context features $C^s$ extracted by PWC-Net from each scale are utilized to construct the weight matrix.
At each position, the feature vector is dot producted with the feature vectors of its neighboring positions (the number of neighboring positions is determined by the fusion range, usually $3\times 3$ or $5\times 5$) to compute the similarities. The results are then normalized using a softmax function.
The weight matrix enables the flow to consider not only the position itself but also its surrounding area, thereby achieving a blurred effect.
\subsection{Helmholtz Velocity Decomposition Constraint}
The deformation of gel materials is complex due to their elastic properties.
Based on the Helmholtz velocity decomposition theorem, compressible motion can be decomposed into four components: translational motion, linear deformation motion, shear deformation motion, and rotational motion, given by
\begin{align}\label{Eq:VD}
{\mathbf{u}}({\mathbf{x}} + \delta {\mathbf{x}}) = {\mathbf{u}}({\mathbf{x}}) + {{\rm X}} \delta {\mathbf{x}} + \Theta \delta {\mathbf{x}} + {\rm Z} \delta {\mathbf{x}},
\end{align}
\begin{align}
{\rm X}  = \left[ {\begin{array}{*{20}{c}}
{{\varepsilon _{xx}}}&0\\
0&{{\varepsilon _{yy}}}
\end{array}} \right],
\Theta  = \left[ {\begin{array}{*{20}{c}}
0&{{\varepsilon _{xy}}}\\
{{\varepsilon _{xy}}}&0
\end{array}} \right],
{\rm Z}  = \left[ {\begin{array}{*{20}{c}}
0&{ - {\omega}}\\
{{\omega}}&0
\end{array}} \right]
\end{align}
where ${\rm X}, {\Theta}$ and ${\rm Z}$ denote the linear distortion rate tensor, shear distortion rate tensor, and rotation tensor, respectively;
$\varepsilon_{xx}=\partial u/ \partial x$ and $\varepsilon_{yy}=\partial v/ \partial y$ are the linear distortion rates in the $x$ and $y$ directions, respectively;
$\varepsilon_{xy}=(\partial u/ \partial y + \partial v/ \partial x)/2$ is the shear distortion rate;
$\omega=(\partial v/ \partial x - \partial u/ \partial y)/2$ is the rotational angular rate.
By decomposing the flow, we can impose more refined constraints on each component, achieving high-precision flow estimation.
Eq. \ref{Eq:VD} can be further transformed as
\begin{equation}\label{Eq:small_motion}
\frac{{{\mathbf{u}}({\mathbf{x}} + \delta {\mathbf{x}}) - {\mathbf{u}}({\mathbf{x}})}}{{\delta {\mathbf{x}}}} = {\rm X}  + {\Theta}  + {\rm Z}.
\end{equation}
Thus, the values of linear distortion rate tensor ${\rm X}$, shear distortion tensor ${\Theta}$, and rotational tensor ${\rm Z}$ are constraint to satisfy the small motions assumption, i.e.
\begin{equation}\label{Eq:VD_sm}
{\left\| {{\rm{vec}}({\rm X})} \right\|_1} + {\lambda _\Theta }{\left\| {{\rm{vec}}(\Theta )} \right\|_1} + {\lambda _{\rm Z} }{\left\| {{\rm{vec}}({\rm Z})} \right\|_1},
\end{equation}
where,
\begin{align}\label{Eq:VD_sm1}
{\left\| {{\rm{vec}}({\rm X} )} \right\|_1} &= |{\varepsilon _{xx}}| + |{\varepsilon _{yy}}|,\\
{\left\| {{\rm{vec}}(\Theta )} \right\|_1} &= 2|{\varepsilon _{xy}}|,\\
{\left\| {{\rm{vec}}({\rm Z})} \right\|_1} &= 2|{\omega}|,
\end{align}
${\lambda _\Theta }$ and ${\lambda _{\rm Z} }$ are coefficients of distortion and rotation tensors, respectively, which values will affect the smooth property of the optical flow;
the function $\rm{vec}(\cdot)$ is used to convert the input into a vector representation.
With the help of the Helmholtz velocity decomposition theorem, the compressible motion can be better estimated \cite{lu_VD}.

Furthermore, the optical flow estimated on the edge of the image is more precise due to the larger gradient.
The deformation of the gel is similar between the edge and other flattened areas. Thus, we adopt an anti-edge-aware weight for the motion smooth term, and the final decomposition loss is written as
\begin{equation}\label{Eq:VD_sm_antiedge}
\mathcal{L}_{dc}^{s} = \left(1 - e^{-\beta \left\| \nabla I_1^s \right\|_1} \right) \left( \left\| \mathrm{vec}(\mathrm{X}) \right\|_1 + \lambda_\Theta \left\| \mathrm{vec}(\Theta) \right\|_1 + \lambda_{\mathrm{Z}} \left\| \mathrm{vec}(\mathrm{Z}) \right\|_1 \right)
\end{equation}
In Eq. \ref{Eq:VD_sm_antiedge}, when an edge is detected (indicated by a larger value in $\left\| {\nabla I_1^s} \right\|_1$), the weight $1 - {e^{ - \beta \left\| {\nabla I_1^s} \right\|_1}}$ increases.
Consequently, the remaining term in $\mathcal{L}_{dc}^s$ must decrease during the minimization process, which enhances the smoothness between the detected edge and the surrounding area.
\begin{figure*}[ht]
  \centering
  \includegraphics[width=0.80\textwidth,clip]{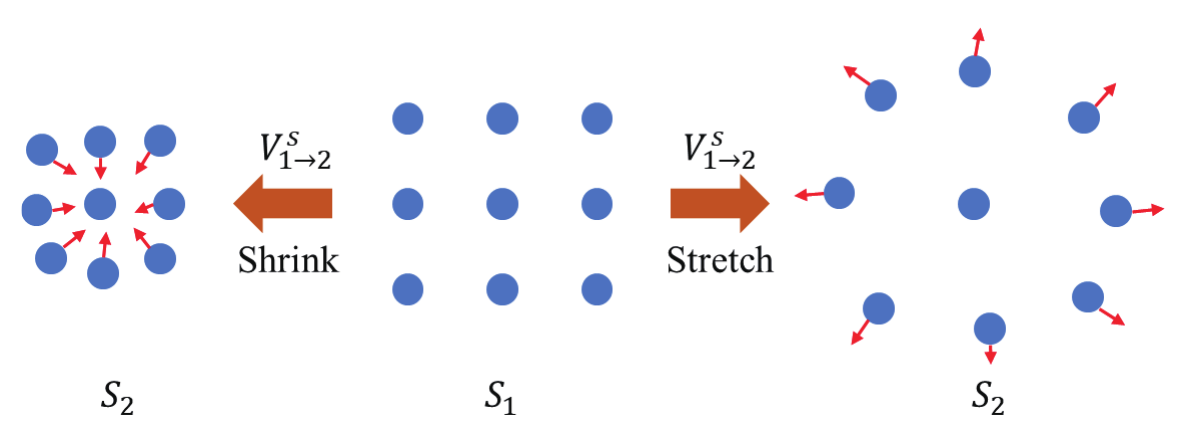}\\
  \caption{Illustration of two typical changes of the gel deformation, with the red vectors indicating the direction of markers' movement.}\label{fig:Expension}
\end{figure*}
\subsection{Elastic Deformation Constraint}
To further constrain the motion change during the elastic deformation of the gel materials, we propose a novel regularization term named the deformation loss term.
As shown in Fig. \ref{fig:Expension}, there are two typical deformations in the motion of gel materials, i.e., shrinking and stretching.
To enforce spatial consistency in gel motion, we can incorporate a constraint that regulates the change in the area between adjacent pixels before and after deformation.
Firstly, the pixel-level area change ratio is estimated between the input image pairs.
Then, the estimated motion is smooth over the entire gel materials by constraining the gradient of the ratio.
Different from \cite{optical_expension}, we calculate the deformation ratio separately for the $x$ and $y$ directions:
\begin{align}\label{Eq:ratio_xy}
({x^{'}} - {x_c^{'}}) &= {\mathcal{R}_x}(x - {x_c}),\quad x \in {\mathcal{N}_{3 \times 3}}({x_c}),\\
({y^{'}} - {y_c^{'}}) &= {\mathcal{R}_y}(y - {y_c}),\quad y \in {\mathcal{N}_{3 \times 3}}({y_c}),
\end{align}
where $x'$ and $y'$ denote the positions $x$ and $y$ warped by the optical flow $V_{1{\rightarrow}2}$;
the subscript $c$ represents the center of the local window;
${\mathcal{N}}_{3 \times 3}$ represents the local window size of $3 \times 3$.
The final deformation ratio is obtained by multiplying the two ratios together:
\begin{align}\label{Eq:ratio}
{\mathcal{R}} = {\mathcal{R}_x}{\mathcal{R}_y}.
\end{align}
Finally, the combined anti-edge-aware weight can be utilized to define the deformation loss as follows:
\begin{align}\label{Eq:df}
\mathcal{L}_{df}^s = (1 - {e^{- \beta {{\left\| {\nabla I_1^s} \right\|}_1}}}){{\left\| {\nabla \mathcal{R}} \right\|}_1}.
\end{align}
\subsection{Loss Function}
The widely used photometric loss function in optical flow tasks is adopted for robust flow estimation, which takes the form:
\begin{equation}\label{Eq:ph}
\mathcal{L}_{ph}^s = \alpha \frac{{1 - \text{SSIM}(\tilde I_1^s, I_1^s)}}{2} + (1 - \alpha ){\left\| {\tilde I_1^s - I_1^s} \right\|_1},
\end{equation}
where SSIM denotes the structural similarity index;
$\alpha$ represents the balance between SSIM and $L_1$ distance;
$\tilde{I}_{1}^{s}$ indicates the warped image $I_{2}^{s}$ using the optical flow $V_{1{\rightarrow}2}^{s}$ at scale $s$.
The photometric loss, combined with the proposed decomposition loss and deformation loss, constructs the loss function at each scale.
The multi-scale loss is defined as the weighted sum of the losses at each scale, denoted by:
\begin{equation}\label{Eq:multi_loss}
\mathcal{L} = \sum\limits_{s = 0}^{l - 2} {{\lambda_s}\mathcal{L}_{self}^s = } \sum\limits_{s = 0}^{l - 2} {{\lambda_s}(\mathcal{L}_{ph}^s + {\lambda_{dc}}\mathcal{L}_{dc}^s + {\lambda_{df}}\mathcal{L}_{df}^s)}, \end{equation}
where $l$ is the number of total scales created by PWC-Net;
$\lambda _{dc}$ and $\lambda _{df}$ are coefficients that control the balance between each loss;
$\lambda _s$ are parameters that weigh the importance of each scale.
\section{Experimental Analysis}\label{sec4}
\subsection{Experiment Setup}\label{sec4sub1}
The proposed self-supervised learning method does not require labeled training data.
Therefore, we extract 1327 image pairs with a resolution of $480\times 640$ pixels from videos captured by \cite{dataset}.
We reserve 8 image pairs with the typical motion of gel deformation (large displacement, shrinking and stretching) for validation and comparison with other optical flow methods.
We train the network for 200 epochs on the training dataset, with a batch size of 4 image pairs per epoch.
Subsequently, we fine-tune the network for an additional 800 epochs using the validation dataset.
The number of pyramid scales, $l$, is set to 8.
The fusion region size of the LFFM is set to $3\times 3$.
In the photometric loss term, $\alpha$ is set to 0.85.
In the decomposition loss term, $\beta$ is set to 10, and both $\lambda_{\Theta}$ and $\lambda_{\rm{Z}}$ are set to 0.01.
In the deformation loss term, $\beta$ is set to 10.
In the multi-scale loss term, $\lambda_s$ is set to 1.0 for each scale, while $\lambda_{dc}$ and $\lambda_{df}$ are set to 75 and 0.01, respectively.
The images are initially resized to a resolution of $512\times 640$ pixels before being fed into the network. The output optical flows are then resized to the original resolution of the images for validation.
\subsection{Evaluation Metrics}\label{sec4sub2}
Since there are no ground truth labels in the dataset, we need to warp the second images into pseudo-first images using the estimated optical flows and compare the similarity between pseudo-first and authentic-first images.
The higher the similarity, the better the estimation.
Two widely used metrics for evaluating image similarity are PSNR (Peak Signal-to-Noise Ratio) and SSIM.
They are defined as follows:
\begin{align}\label{PSNR_SSIM}
\text{PSNR}(I,\tilde I) &= 10 \times {\log _{10}}\left( {\frac{{{{({2^n} - 1)}^2}}}{{\text{MSE}(I,\tilde I)}}} \right),\\
\text{SSIM}(I,\tilde I) &= \frac{{(2{\mu _x}{\mu _y} + {c_1})(2{\sigma _{xy}} + {c_2})}}{{(\mu _x^2 + \mu _y^2 + {c_1})(\sigma _x^2 + \sigma _y^2 + {c_2})}},
\end{align}
where, $n$ represents the bit depth of the pixels;
$\text{MSE}(I, \tilde{I})$ is the mean square error between the input image $I$ and the warped image $\tilde{I}$;
$\mu_x$ and $\mu_y$ are the means of $I$ and $\tilde{I}$, respectively;
$\sigma_x$ and $\sigma_y$ are the variances of $I$ and $\tilde{I}$, and $\sigma_{xy}$ represents their covariance;
$c_1$ and $c_2$ are constants used to maintain stability.
Therefore, we will use these two metrics for comparisons and evaluations in the following subsections.
\begin{table}[]
\centering
\newcolumntype{Z}{>{\centering\arraybackslash}X}
\caption{Comparison with traditional and deep learning-based optical flow methods using the validation dataset. '\#' represents an image pair. The best and the second-best values within each category are marked as bold and underlined, respectively. The best value among all the methods is marked in red. 'ft' denotes fine-tuning the model on the validation dataset.}\label{tab:Methods}
\begin{tabularx}{\textwidth}{p{0.4cm}lp{0.2cm}cp{0.2cm}XXXXXXXX}
\toprule
\multicolumn{1}{l}{}            & \multicolumn{1}{c}{Method}        &  & Metric &  & \#1   & \#2   & \#3   & \#4   & \#5   & \#6   & \#7   & \#8   \\ \midrule
\multirow{10}{*}{\rotatebox{90}{Traditional}}   & \multirow{2}{*}{Farneback}        &  & PSNR   &  & 38.48 & 35.02 & 32.57 & 32.03 & 31.73 & 32.31 & 33.70 & 32.20 \\ \cline{3-13}
                                &                                   &  & SSIM   &  & 0.96  & 0.91  & 0.91  & 0.89  & 0.89  & 0.91  & 0.91  & 0.90  \\ \cline{2-13}
                                & \multirow{2}{*}{DIS (Ultra-fast)} &  & PSNR   &  & 39.64 & 35.12 & 32.89 & 33.23 & 32.63 & 32.92 & 34.93 & 32.39 \\ \cline{3-13}
                                &                                   &  & SSIM   &  & 0.97  & 0.92  & 0.92  & 0.92  & 0.92  & 0.93  & 0.94  & 0.91  \\ \cline{2-13}
                                & \multirow{2}{*}{DIS (Fast)}       &  & PSNR   &  & 39.75 & 35.26 & 32.89 & 33.28 & 32.60 & 32.97 & 35.11 & 32.49 \\ \cline{3-13}
                                &                                   &  & SSIM   &  & 0.97  & 0.92  & 0.92  & 0.92  & 0.92  & 0.93  & 0.94  & 0.91  \\ \cline{2-13}
                                & \multirow{2}{*}{DIS (medium)}     &  & PSNR   &  & $\mathbf{40.25}$ & $\underline{35.41}$ & $\underline{33.25}$ & $\underline{33.86}$ & $\underline{33.27}$ & $\underline{33.24}$ & $\underline{35.41}$ & $\underline{32.73}$ \\ \cline{3-13}
                                &                                   &  & SSIM   &  & $\mathbf{0.97}$  & $\mathbf{0.92}$  & $\mathbf{0.92}$  & $\mathbf{0.93}$  & $\mathbf{0.93}$  & $\mathbf{0.93}$  & $\mathbf{0.94}$  & $\mathbf{0.91}$  \\ \cline{2-13}
                                & \multirow{2}{*}{TV-L1}            &  & PSNR   &  & $\underline{39.98}$ & $\mathbf{35.44}$ & $\mathbf{33.31}$ & $\mathbf{33.95}$ & $\mathbf{33.68}$ & $\mathbf{33.42}$ & $\mathbf{35.64}$ & $\mathbf{32.88}$ \\ \cline{3-13}
                                &                                   &  & SSIM   &  & $\underline{0.97}$  & $\underline{0.92}$  & $\underline{0.92}$  & $\underline{0.92}$  & $\underline{0.93}$  & $\underline{0.93}$  & $\underline{0.94}$  & $\underline{0.91}$  \\ \hline
\multirow{12}{*}{\rotatebox{90}{Deep Learning}} & \multirow{2}{*}{RAFT}             &  & PSNR   &  & 37.73 & 34.83 & 31.49 & 31.11 & 31.17 & 32.44 & 34.78 & 31.57 \\ \cline{3-13}
                                &                                   &  & SSIM   &  & 0.97  & 0.91  & 0.91  & 0.91  & 0.91  & 0.92  & 0.94  & 0.90  \\ \cline{2-13}
                                & \multirow{2}{*}{ARFlow}           &  & PSNR   &  & 39.76 & 35.22 & 32.88 & 33.44 & 32.54 & 33.05 & 35.30 & 32.46 \\ \cline{3-13}
                                &                                   &  & SSIM   &  & 0.97  & 0.92  & 0.92  & 0.92  & 0.92  & 0.93  & 0.94  & 0.91  \\ \cline{2-13}
                                & \multirow{2}{*}{SelfFlow}           &  & PSNR   &  & 40.22 & 35.31 & 33.12 & 33.78 & 32.99 & 33.62 & 35.44 & 32.65 \\ \cline{3-13}
                                &                                   &  & SSIM   &  & 0.97  & 0.92  & 0.92  & 0.93  & 0.93  & 0.93  & 0.94  & 0.91  \\ \cline{2-13}
                                & \multirow{2}{*}{SelfFlow+ft}        &  & PSNR   &  & 40.36 & 35.74 & 33.35 & 34.30 & 33.64 & 34.22 & 35.90 & 33.02 \\ \cline{3-13}
                                &                                   &  & SSIM   &  & 0.97  & 0.93  & 0.92  & 0.93  & 0.93  & 0.94  & 0.95  & 0.92  \\ \cline{2-13}
                                & \multirow{2}{*}{GelFlow}          &  & PSNR   &  & $\underline{40.57}$ & $\underline{35.90}$ & $\underline{33.43}$ & $\underline{34.34}$ & $\underline{33.65}$ & $\underline{33.96}$ & $\underline{35.94}$ & $\underline{33.11}$ \\ \cline{3-13}
                                &                                   &  & SSIM   &  & $\underline{0.98}$  & $\underline{0.93}$  & $\underline{0.93}$  & $\underline{0.93}$  & $\underline{0.93}$  & $\underline{0.94}$  & $\underline{0.95}$  & $\underline{0.93}$  \\ \cline{2-13}
                                & \multirow{2}{*}{GelFlow+ft}       &  & PSNR   &  & \textcolor{red}{$\mathbf{40.76}$} & \textcolor{red}{$\mathbf{36.00}$} & \textcolor{red}{$\mathbf{33.66}$} & \textcolor{red}{$\mathbf{34.71}$} & \textcolor{red}{$\mathbf{34.25}$} & \textcolor{red}{$\mathbf{35.00}$} & \textcolor{red}{$\mathbf{36.13}$} & \textcolor{red}{$\mathbf{33.22}$} \\ \cline{3-13}
                                &                                   &  & SSIM   &  & \textcolor{red}{$\mathbf{0.98}$}  & \textcolor{red}{$\mathbf{0.93}$}  & \textcolor{red}{$\mathbf{0.93}$}  & \textcolor{red}{$\mathbf{0.94}$}  & \textcolor{red}{$\mathbf{0.94}$}  & \textcolor{red}{$\mathbf{0.95}$}  & \textcolor{red}{$\mathbf{0.95}$}  & \textcolor{red}{$\mathbf{0.93}$}  \\ \bottomrule
\end{tabularx}
\end{table}
\subsection{Comparisons with Classical Optical Flow Methods}\label{sec4sub3}
We compared traditional dense optical flow methods (Farnaback, DIS, TV-L1) using OpenCV and deep learning-based optical flow methods (RAFT \cite{RAFT}, ARFlow \cite{ARFlow}, and a self-supervised method named SelfFlow using a photometric loss mentioned before with a first-order smoothness loss \cite{UFlow}).
The results of the comparison are presented in Table \ref{tab:Methods}.
Notably, the self-supervised methods SelfFlow and GelFlow were fine-tuned using the strategy described in Section \ref{sec4sub1}.
The validation dataset showed that TV-L1 and DIS (medium) performed similarly and outperformed other traditional methods. However, the solving strategy of TV-L1 is time-consuming, making it much slower than the optimized DIS method.
Consequently, the DIS method is widely used as the dense optical flow estimator in current vision-based tactile sensors.

It is worth mentioning that we directly utilized the pre-trained models of RAFT and ARFlow, testing them on the vision-based tactile dataset.
Therefore, their performance may not be satisfactory.
On the other hand, SelfFlow and GelFlow were trained on the dataset and further fine-tuned.
As a result, they outperformed the existing traditional methods.
The excellent performance can be attributed to the strong learning ability of convolutional neural networks and the well-designed loss functions guiding the network output towards the ground truth.
Among all the candidate methods, GelFlow achieved the best performance with its proposed flow fusion operation and motion decomposition and deformation loss, which guide the parameters of the network towards global optimization.
In conclusion, the comparisons indicate that the proposed GelFlow method is particularly adept at handling gel materials' deformation.
\section{Conclusion}\label{sec5}
In this study, we propose the GelFlow method, which incorporates several novel components to address the challenges posed by gel deformation.
Firstly, GelFlow constructs a multi-scale feature pyramid to extract hidden features from the input image pairs and handle large displacements effectively.
A local flow fusion module fuses the flow using neighboring flows with appropriate weights.
This fusion process achieves a blurred effect, which is crucial for capturing the deformations occurring in gel materials.
We propose two novel loss functions to better handle the intricate gel deformations: the velocity decomposition loss and the elastic deformation loss.
A photometric loss combined with the proposed two novel motion smoothness losses is used to construct the multi-scale loss to better guide the network from global optimization.
Finally, the network is trained in a self-supervised manner, and the comparison result with other optical flow methods indicates that the GelFlow method performs the best due to the superior capacity of the convolutional neural networks to extract valuable features and the strong ability of global optimization.
%
%
%
\bibliographystyle{GelFlow}
\bibliography{GelFlow}

\begin{thebibliography}{10}
\providecommand{\url}[1]{\texttt{#1}}
\providecommand{\urlprefix}{URL }
\providecommand{\doi}[1]{https://doi.org/#1}

\bibitem{introduction_1}
Liu, H., Yu, Y., Sun, F., Gu, J.: Visual--tactile fusion for object
  recognition. IEEE Transactions on Automation Science and Engineering
  \textbf{14}(2),  996--1008 (2016)

\bibitem{introduction_3}
Zhang, S., Chen, Z., Gao, Y., Wan, W., Shan, J., Xue, H., Sun, F., Yang, Y.,
  Fang, B.: Hardware technology of vision-based tactile sensor: A review. IEEE
  Sensors Journal  (2022)

\bibitem{GelForce}
Sato, K., Kamiyama, K., Kawakami, N., Tachi, S.: Finger-shaped gelforce: sensor
  for measuring surface traction fields for robotic hand. IEEE Transactions on
  Haptics  \textbf{3}(1),  37--47 (2009)

\bibitem{geometry}
Cui, S., Wang, R., Hu, J., Zhang, C., Chen, L., Wang, S.: Self-supervised
  contact geometry learning by gelstereo visuotactile sensing. IEEE
  Transactions on Instrumentation and Measurement  \textbf{71}, ~1--9 (2021)

\bibitem{slip}
James, J.W., Lepora, N.F.: Slip detection for grasp stabilization with a
  multifingered tactile robot hand. IEEE Transactions on Robotics
  \textbf{37}(2),  506--519 (2020)

\bibitem{random_color_pattern}
Du, Y., Zhang, G., Zhang, Y., Wang, M.Y.: High-resolution 3-dimensional contact
  deformation tracking for fingervision sensor with dense random color pattern.
  IEEE Robotics and Automation Letters  \textbf{6}(2),  2147--2154 (2021)

\bibitem{OpenCV}
Bradski, G.: The opencv library. Dr. Dobb's Journal: Software Tools for the
  Professional Programmer  \textbf{25}(11),  120--123 (2000)

\bibitem{Deltact}
Zhang, G., Du, Y., Yu, H., Wang, M.Y.: Deltact: A vision-based tactile sensor
  using a dense color pattern. IEEE Robotics and Automation Letters
  \textbf{7}(4),  10778--10785 (2022)

\bibitem{GelSight}
Yuan, W., Dong, S., Adelson, E.H.: Gelsight: High-resolution robot tactile
  sensors for estimating geometry and force. Sensors  \textbf{17}(12), ~2762
  (2017)

\bibitem{GelSlim}
Donlon, E., Dong, S., Liu, M., Li, J., Adelson, E., Rodriguez, A.: Gelslim: A
  high-resolution, compact, robust, and calibrated tactile-sensing finger. In:
  2018 IEEE/RSJ International Conference on Intelligent Robots and Systems
  (IROS). pp. 1927--1934. IEEE (2018)

\bibitem{DIGIT}
Lambeta, M., Chou, P.W., Tian, S., Yang, B., Maloon, B., Most, V.R., Stroud,
  D., Santos, R., Byagowi, A., Kammerer, G., et~al.: Digit: A novel design for
  a low-cost compact high-resolution tactile sensor with application to in-hand
  manipulation. IEEE Robotics and Automation Letters  \textbf{5}(3),
  3838--3845 (2020)

\bibitem{DenseTact}
Do, W.K., Kennedy, M.: Densetact: Optical tactile sensor for dense shape
  reconstruction. In: 2022 International Conference on Robotics and Automation
  (ICRA). pp. 6188--6194. IEEE (2022)

\bibitem{OmniTact}
Padmanabha, A., Ebert, F., Tian, S., Calandra, R., Finn, C., Levine, S.:
  Omnitact: A multi-directional high-resolution touch sensor. In: 2020 IEEE
  International Conference on Robotics and Automation (ICRA). pp. 618--624.
  IEEE (2020)

\bibitem{GelStereo}
Cui, S., Wang, R., Hu, J., Wei, J., Wang, S., Lou, Z.: In-hand object
  localization using a novel high-resolution visuotactile sensor. IEEE
  Transactions on Industrial Electronics  \textbf{69}(6),  6015--6025 (2021)

\bibitem{Tac3D}
Zhang, L., Wang, Y., Jiang, Y.: Tac3d: A novel vision-based tactile sensor for
  measuring forces distribution and estimating friction coefficient
  distribution. arXiv preprint arXiv:2202.06211  (2022)

\bibitem{LK}
Bouguet, J.Y., et~al.: Pyramidal implementation of the affine lucas kanade
  feature tracker description of the algorithm. Intel corporation
  \textbf{5}(1-10), ~4 (2001)

\bibitem{FB}
Farneb{\"a}ck, G.: Two-frame motion estimation based on polynomial expansion.
  In: Image Analysis: 13th Scandinavian Conference, SCIA 2003 Halmstad, Sweden,
  June 29--July 2, 2003 Proceedings 13. pp. 363--370. Springer (2003)

\bibitem{DIS}
Kroeger, T., Timofte, R., Dai, D., Van~Gool, L.: Fast optical flow using dense
  inverse search. In: Computer Vision--ECCV 2016: 14th European Conference,
  Amsterdam, The Netherlands, October 11--14, 2016, Proceedings, Part IV 14.
  pp. 471--488. Springer (2016)

\bibitem{Soft_bubble}
Kuppuswamy, N., Alspach, A., Uttamchandani, A., Creasey, S., Ikeda, T.,
  Tedrake, R.: Soft-bubble grippers for robust and perceptive manipulation. In:
  2020 IEEE/RSJ International Conference on Intelligent Robots and Systems
  (IROS). pp. 9917--9924. IEEE (2020)

\bibitem{sun2018pwc}
Sun, D., Yang, X., Liu, M.Y., Kautz, J.: Pwc-net: Cnns for optical flow using
  pyramid, warping, and cost volume. In: Proceedings of the IEEE conference on
  computer vision and pattern recognition. pp. 8934--8943 (2018)

\bibitem{warping}
Brox, T., Bruhn, A., Papenberg, N., Weickert, J.: High accuracy optical flow
  estimation based on a theory for warping. In: European conference on computer
  vision. pp. 25--36. Springer (2004)

\bibitem{TactileFlow}
Du, Y., Zhang, G., Wang, M.Y.: 3d contact point cloud reconstruction from
  vision-based tactile flow. IEEE Robotics and Automation Letters
  \textbf{7}(4),  12177--12184 (2022)

\bibitem{lu_VD}
Lu, J., Yang, H., Zhang, Q., Yin, Z.: An accurate optical flow estimation of
  piv using fluid velocity decomposition. Experiments in Fluids  \textbf{62},
  1--16 (2021)

\bibitem{optical_expension}
Yang, G., Ramanan, D.: Upgrading optical flow to 3d scene flow through optical
  expansion. In: Proceedings of the IEEE/CVF Conference on Computer Vision and
  Pattern Recognition. pp. 1334--1343 (2020)

\bibitem{dataset}
Yang, F., Ma, C., Zhang, J., Zhu, J., Yuan, W., Owens, A.: Touch and go:
  Learning from human-collected vision and touch. In: Thirty-sixth Conference
  on Neural Information Processing Systems Datasets and Benchmarks Track
  (2022), \url{https://openreview.net/forum?id=ZZ3FeSSPPblo}

\bibitem{RAFT}
Teed, Z., Deng, J.: Raft: Recurrent all-pairs field transforms for optical
  flow. In: Computer Vision--ECCV 2020: 16th European Conference, Glasgow, UK,
  August 23--28, 2020, Proceedings, Part II 16. pp. 402--419. Springer (2020)

\bibitem{ARFlow}
Liu, L., Zhang, J., He, R., Liu, Y., Wang, Y., Tai, Y., Luo, D., Wang, C., Li,
  J., Huang, F.: Learning by analogy: Reliable supervision from transformations
  for unsupervised optical flow estimation. In: IEEE Conference on Computer
  Vision and Pattern Recognition(CVPR) (2020)

\bibitem{UFlow}
Jonschkowski, R., Stone, A., Barron, J.T., Gordon, A., Konolige, K., Angelova,
  A.: What matters in unsupervised optical flow. In: Computer Vision--ECCV
  2020: 16th European Conference, Glasgow, UK, August 23--28, 2020,
  Proceedings, Part II 16. pp. 557--572. Springer (2020)

\end{thebibliography}
\end{document}